# Semantic maps and metrics for science using deep transformer encoders

**Brendan Chambers**
QuillBot
Knowledge Lab at the University of Chicago

**James Evans**
Department of Sociology and Knowledge Lab, University of Chicago
Santa Fe Institute

*Corresponding author*
Brendan Chambers, brendan.chambers@quillbot.com
James Evans, jevans@uchicago.edu

*Affiliation address*
Knowledge Lab
University of Chicago
1155 E 60th St | Room 211
Chicago, IL 60637

# Semantic maps and metrics of science using deep transformer encoders

## ABSTRACT

The growing deluge of scientific publications demands text analysis tools that can help scientists and policy-makers navigate, forecast and beneficially guide scientific research. Recent advances in natural language understanding driven by deep transformer networks offer new possibilities for mapping science. Because the same surface text can take on multiple and sometimes contradictory specialized senses across distinct research communities, sensitivity to context is critical for infometric applications. Transformer embedding models such as BERT capture shades of association and connotation that vary across the different linguistic contexts of any particular word or span of text. Here we report a procedure for encoding scientific documents with these tools, measuring their improvement over static word embeddings in a nearest-neighbor retrieval task. We find discriminability of contextual representations is strongly influenced by choice of pooling strategy for summarizing the high-dimensional network activations. Importantly, we note that fundamentals such as domain-matched training data are more important than state-of-the-art NLP tools. Yet state-of-the-art models did offer significant gains. The best approach we investigated combined domain-matched pretraining, sound pooling, and state-of-the-art deep transformer network encoders. Finally, with the goal of leveraging contextual representations from deep encoders, we present a range of measurements for understanding and forecasting research communities in science.

## HIGHLIGHTS

Pretrained transformer network encoders such as *BERT* have become the champion architecture for modeling contextual representations in text data.

These models benefit greatly from fine-tuning on labeled datasets, so there is ongoing discussion as to whether they are suitable for unsupervised embeddings.

We develop and validate a pipeline for mapping content across scientific publications using an unsupervised *sciBERT* encoder.

Two key design considerations for practitioners are highlighted: pretraining history, and pooling strategy.

We also provide a streamlined parser for acquiring PubMed tile/abstract text and writing to a relational database.

## 1.1
## INTRODUCTION

The same structural features of scientific research that mark its institutional ascendance—specialization and jargon, field and subfield societies and governance, and vast volumes of published work—also pose challenges for the transmission of knowledge. Whether for researchers or the general public, it is harder than ever to understand, navigate and beneficially guide the landscape of concurrent research paradigms in complex relation to one another.

Prior efforts to leverage text analysis for the comprehension and summary of scientific research have made substantial progress over the past three decades. These include topic modeling, latent semantic analysis and word embedding models. Topic modeling is based on a Bayesian probability model that inductively discovers "topics" structuring a corpus, each learned as a sparse distribution over words that co-occur in text (Blei et al., 2003; Zeng et al., 2012). Latent Semantic Analysis (LSA), an early approach to word embedding, used singular-value decomposition (SVD) to factorize a word-by-document matrix (Dumais, 2004). Neural network word embeddings increased the power of these embedding methods through improved representation and optimization strategies, yet likely perform a similar function of approximately factorizing word and context matrices (Levy & Goldberg, 2014). This enabled analysts to consider only the local, most relevant contexts for representing words as vectors in a shared vector space, such that words sharing similar local, linguistic contexts will be positioned nearby in the space and those that appear only in distinct and disconnected contexts will be positioned apart.

Technical advances in natural language processing—namely, rapid improvements in pretrained deep neural networks (Dai & Le, 2015; Devlin et al., 2018; Howard & Ruder, 2018; Peters et al., 2018; Radford et al., n.d.; Vaswani et al., 2017)—lend increased sensitivity and power to text based approaches. Most importantly, new technologies are sensitive to lexical composition (Shwartz & Dagan, 2019) and contextual semantics (Ethayarajh, 2019). That is, the vector representation of a single word will vary according to its context in the entire abstract. Across all abstracts, a single word will occupy a distribution of points with one or more clusters: its vector representation in the latent space will incorporate substructure and contextual variation from its distributional statistics. In other words, across the corpus, a word is not represented by a single vector indexing one point in high-dimensional meaning space. Rather, each word is represented by a cloud of points, one vector for each occurrence, tracing the word's varying meanings. For analysis of scientific writing, where a single word can take on multiple and sometimes contradictory specialized senses, especially across the boundaries that separate research communities (Gieryn, 1999), sensitivity to context is an important methodological development.

Without minimizing the importance of these developments, it is also important to recognize that language models lag behind human readers in many important ways. For example, early hype around common-sense reasoning performance was later moderated, with the recognition of spurious statistical cues in benchmark datasets (Gururangan et al., 2018; Niven & Kao, 2019; Thomas McCoy et al., 2019). Indeed, evaluation with controlled probes reveals limitations in common pretrained models (Ettinger, 2020). Deep transformer models are also polluted by social biases in human cognition (Caliskan & Lewis, 2020) including racism, sexism, regressive parsing of gender, and harmful stereotypes about disabilities (Birhane & Guest, 2020; Hutchinson et al., 2020; Molly Lewis & Lupyan, 2020; May et al., 2019). These harms manifest intersectionally (Guo & Caliskan, 2020) and they cannot be simply transformed away. Though imperfect, mitigation strategies are crucial (Bhardwaj et al., 2020; Sun et al., 2019) and need to be designed around a given context of deployment. There is also need for more foundational solutions, beyond downstream technical mitigations (Birhane & Cummins, 2019; Gebru et al., 2018; Jo & Gebru, 2020). Because these systems are in widespread use, there is a pressing need for further work in characterizing and mitigating their harms (Blodgett et al., 2020; Gehman et al., 2020; Suresh & Guttag, 2019).

In this work, we develop and validate a procedure for encoding scientific abstracts with deep transformer encoders, improving over a baseline based on static, shallow neural network word embeddings. Importantly, we find that fundamentals such as domain-matched training data are more important than state-of-the-art NLP tools. Similarly, we find that a given pooling strategy for summarizing information from deep transformer networks can make the difference between low- versus high-performance. Finally, we find that the best approach of all combines domain-appropriate training, sound pooling, and state-of-the-art deep transformer network encoders. Under these conditions, our results demonstrate that transformer network encoders (Q. Liu et al., 2020; Rogers et al., 2020) are a powerful option for organizing scientific abstracts according to similarities in their linguistic conventions and semantic content.

## 1.2
## MATERIALS AND METHODS

### 1.2.1
*PubMed data*

The PubMed collection is a resource created by the National Center for Biotechnology Information, part of the U.S. National Library of Medicine at the National Institutes of Health (NIH). PubMed contains more than 30 million citations drawn from the MEDLINE collection, life science journals, and online books. It spans biomedicine and health, including overlapping contributions from related sciences and is a free, open resource. Thanks to excellent data engineering support at NIH, bulk downloads are available. They are hosted at the following url: https://www.nlm.nih.gov/databases/download/pubmed_medline.html

### 1.2.2
*Computing resources*

This project was made possible thanks to the Midway2 computing cluster developed and maintained by the Research Computing Center at the University of Chicago.

### 1.2.3
*Bulk download and XML parsing*

For each of the citations in the PubMed dataset, we parsed PMID, title, abstract, publication journal, and publication year. In some cases, particularly for older records, abstract text was not available. Parsing was accomplished with a custom pipeline created by the authors. First, compressed XML documents were obtained from the NIH FTP servers named above. XML files were parsed using the LXML python parser, organized into relational tables, and written to a custom MySQL database. Operations were orchestrated in parallel using Pyspark, running in local mode. Communication between Pyspark and the MySQL database was coordinated using the Java Database Connector (JDBC). Code is is available at the following URL: https://github.com/brendanchambers/parse_pubmed

1.2.4
*Text preprocessing*
Title and abstract text were concatenated with a separating period mark. Text was transformed to lower case. Numbers were replaced with a mask token <NUM>.

1.2.5
*Transformer encoder models*
Transformer encoder models were implemented through the excellent HuggingFace library (Wolf et al., 2019), using the checkpoints *bert-base-uncased (Devlin et al., 2018)* and *scibert-scivocab-uncased (Beltagy et al., 2019)*.

1.2.6
*Encoding semantic features of text*
For entries with both title and abstract text available, the two fields were concatenated prior to encoding. Text was processed through each of the three encoder models enumerated above. Embeddings for entries with title field only appeared similar to entries with significantly longer text, where both title and abstract were present. However, since their norms were expected to be smaller (based e.g. on a random walk baseline for token pooling), entries without abstracts might be trivially distinguished from entries that did have abstracts. In principle, normalization of token vectors and/or pooled vectors will correct for these differences in input text length; but best practices for normalization of transformer activations is not yet established. Therefore, below, we analyze only those entries with data for both *title* and *abstract* fields.

We used the popular algorithm word2vec as a source of static word embeddings (Mikolov et al., 2013). The word2vec algorithm has a single hidden layer and was partially responsible for a wave of widespread interest in the power of vector representations for capturing compositional aspects of language. For the word2vec-pubmed model, pre-trained word vectors were retrieved for each word in the input title and abstract, normalized to the unit hyper-sphere (so that each word vector has the length measured according to the L2 norm), and mean-pooled (sum each word and divide by the number of words in the input text), resulting in a 300-dimensional summary vector for each encoded article. Out of vocabulary words were omitted. The pretrained word2vec model was originally implemented using the skip-gram formulation under hierarchical softmax training, a window size of 5, and subsampling of frequent words (threshold 0.001).

For the pretrained BERT and sci-BERT models, text was first tokenized and truncated if necessary, then encoded using the transformer models. Top-layer activations were pooled as a mean over all non-empty input tokens, omitting the <CLS> and <SEP> tokens, resulting in a 768-dimensional summary vector for each encoded article. For experiments comparing representations for individual tokens, the spacy-transformers library was used to align subword tokens to full words and input text to top-layer activations.

### 1.2.7
*Qualitative analysis of contextual word representations*

A set of sample text was tokenized (title and abstract concatenations, $n$=100), establishing an estimate of the most-frequent subword tokens. Probe words were selected from within this set of common tokens. Recording every top-layer activation of a transformer model is relatively expensive in memory. Choosing frequent tokens enables robust sampling from a minimal number of abstracts.

Probe words were analyzed as contextual embedding vectors, based on their corresponding top-layer hidden-state activations. To perform this analysis, a second set of sample text was tokenized and encoded (title and abstract concatenations, $n$=1000). For tokens appearing in the probe set, top-layer activations were recorded. Recorded activations were visualized using principal component analysis.

### 1.2.8
*Post-processing and comparison of encoder models*

Representations for the three models were post-processed using a random sample of articles ($n$=100000). For each model, vector representations of abstracts were observed to be strongly anisotropic, as expected based on other reports (Ethayarajh, 2019; Zhou et al., 2019). For each model, overall mean encodings were estimated from the sample set. Representations were demeaned by subtracting this shared component from all articles.

Dimensionality reduction was then performed to (1) enable fair comparison of the word2vec versus transformer network encodings, where dimensionality was matched, (2) enable estimation of densities, and (3) facilitate visualization.

### 1.2.9
*Dimensionality reduction in figures and analysis*

For visualization, contextual word vectors were transformed from their original size of 768-dimensions down to 2-dimensions, using principal component analysis (*sklearn.decomposition.PCA*).

Probe words across different output channels were transformed from their original size of 768-dimensions down to 2-dimensions, using *umap* (*umap-learn)*. Compared to *PCA*, *umap* is a more aggressive dimensionality reduction method, permitting non-linear transformations that preserve nearest-neighbor relationships. From a theoretical perspective, this is motivated by the desire to recover lower dimensional manifold structure contained within the original high-dimensional samples (McInnes et al., 2018).

Pooled representations of title-abstract text were transformed using PCA as above, for a sample consisting of all entries from *Journal of Neurophysiology* or *NeuroImage*.

Published title-abstracts and cited title-abstracts were transformed from 768-dimensions to 2-dimensions with *umap*. In order to establish a map with historical context, title-abstracts were sampled randomly from the entire PubMed/MEDLINE database (*n=100,000*). This reference set was used to train a *umap* transformation. The set of publications and references from 2009 was then projected into the trained *umap* coordinates, based on the nearest neighbor interpolation scheme implemented in *umap-learn*.

For benchmarking, pooled transformer representations of title-abstract text were transformed from 768-dimensions to 100-dimensions, to permit fair comparison in this study, as well as future comparison with models having as few as 100 dimensions. Similarly, pooled word2vec representations were transformed from 300-dimensions to 100-dimensions.

1.2.10
*Comparison of model choice and pooling strategies*

A representative proxy task for information science applications was assessed, drawing on the authors' expert knowledge of neuroscience to define a challenging discriminability application. Two journals were chosen, serving largely distinct research communities, yet sharing a similar specialty vocabulary (albeit employed with differing connotations). Pooled representations were assessed to determine whether they could differentiate among these two journals, drawn from distinct nearby specialty areas. Sample title-abstract strings were drawn from each journal, encoded to embedding coordinates in various ways for comparison. Samples were transformed into estimated densities for visualization, estimated using a kernel density method (*seaborn kde*).

To quantify the utility of each embedding space, we assessed the discriminability of their encodings in this challenging setting: two nearby yet distinct specialty journals with different core areas of interest, different breadth of topics, yet overlapping specialty vocabularies (see *Methods*). Metrics of performance needed to (1) operate on the raw embedding space and (2) assess both precision and recall. Embedding spaces were matched in dimensionality using *PCA*, with all evaluations conducted at *D*=100.

With a view towards typical science-of-science use cases, we first consider the problem of finding the 500-most similar articles to a given entry, restricted to the subset of data belonging to either of the two journals considered for evaluation. That is, for the construction of the proxy task, these articles could be drawn from either *Journal of Neurophysiology* or *NeuroImage*. *NeuroImage* serves a multi-disciplinary audience centered on imaging neuroscience, while *Journal of Neurophysiology* emphasizes articles about the function of the nervous system--a subtle distinction for outsiders, but one which is salient to specialists. Since these journals serve distinct research communities, a strong encoder should rate abstract pairs drawn from a single journal as more similar on average than pairs drawn from opposite journals. For a sample title-abstract encoding from journal *J* in a given embedding subspace, we assess the fraction of its 500 nearest neighbors belonging to the same journal *J* (*precision@500*). Nearest neighbors were assessed under the L2 metric. This measurement was aggregated over 5000 samples, reporting mean precision among the 500 nearest neighbors.

We extend this evaluation for two less typical but more comprehensive cases. First, for a sample from journal *J*, let the total number of samples from journal *J* be *R*. We assess the fraction of the *R* nearest neighbors drawn from the same journal (*precision@R*), mean aggregated over 5000 samples. Finally, to differentiate between potential performance differences at all recall cutoffs, we report mean average precision at *R* (*MAP@R*), aggregated over 5000 samples.

1.2.11
*Graphics*

Figures were generated using *matplotlib.pyplot* and *seaborn* python packages in *svg* format. Axis text, titles, legends, colors, and transparency were post-processed using *Inkscape* and *Adobe Illustrator.*

# 1.3
# RESULTS

### 1.3.1
*Data*

First, the entirety of the biomedical research collection *PubMed* was obtained from the U.S. National Center for Biotechnology Information at the National Institute of Health (NIH). *PubMed* is a large research corpus curated to encompass the breadth of biomedical research, with permissive licensing and daily updates (see *Methods* for more information). *PubMed* is available for bulk download via NIH FTP servers, packaged as compressed *xml* files containing rich data for each constituent publication.

Within the larger plethora of available data, a small subset of fields were relevant to this work: PubMed identifier PMID, title text, abstract text, publication journal, and publication year were parsed from the compressed *xml* files. Abstract text was not available for all entries, particularly for older historical data. In this work, entries without abstracts were simply omitted from the encoding analysis.

To extract fields from within the *xml* data, custom parsing was implemented using the *lxml* parser, executed in parallel using *PySpark* in local mode (see *Methods)*. Parsed data were inserted into a MySQL database, indexed by journal and year, to facilitate access in downstream analysis. Next, text was encoded using two transformer network encoders (*BERT* and *sci-BERT*) and a domain-specific word2vec model (see *Methods*).

### 1.3.2
*Contextual embeddings reflect semantics for individual words*

Static word embeddings may be more familiar to the reader than contextual embeddings, for which vector representation coordinates differ from one inference context to another (e.g. in different sentences). Before implementing a procedure to embed full text (title and abstract) we verified that representations for individual words co-located similar (and differentiate dissimilar) tokens.

The tokens *%* and *percent* are very different in their orthographic forms, but very similar semantically. We verified that the sci-BERT model projects these two tokens to nearby coordinates. Using *umap* to visualize the space of variation contained within our sample, *%* and *percent* occupied nearly indistinguishable areas. Their variability across sampled contexts was small compared to the full space (Figure 1a).

The tokens *human* and *cells* are both used broadly across biomedical research areas, in varied contexts. The two words are associated with fundamentally distinct scales and distinct semantics, yet their large degree of contextual variation might be expected to blur their representations together, risking conflating their semantics. We verified that *human* and *cell* exhibited contextual sensitivity while maintaining their discriminability, judged based on qualitative separability (Figure 1b). Principal components analysis was used to visualize the set of sampled tokens in two dimensions. The tokens *human* and *cells* occupied opposite extremes of the sampled representation space.

Because transformer network encoders process a sequence of text, they produce multiple outputs (top-layer activations)—one for each input token. We verify that these outputs from parallel channels can be sensibly combined. That is, we verify that parallel contextual output channels are aligned, in a shared representation space. Because embedding-layer weights are tied at the input and output level and shared across channels, alignment of output channels is not expected to be an issue. We verify this empirically with a set of probe words, comparing their contextual representations when they appear at the first versus second sequence positions (i.e. channels). We present a larger sample, visualized with *umap*, qualitatively confirming that probe words maintain stable representation coordinates across the two sequence positions (Figure 1c).

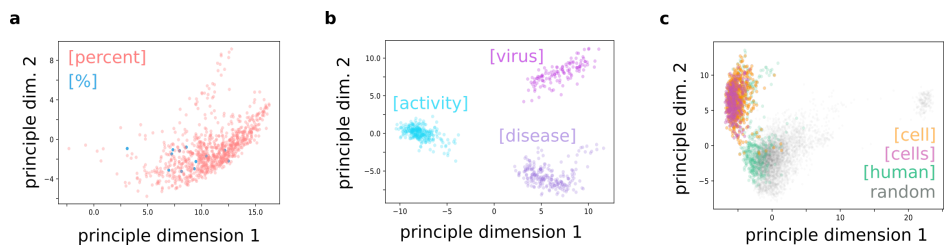

**Figure 1** Contextual word representations from sci-BERT
(a) The token *%* and the token *percent* co-locate (marked with the symbol *x*) within a *umap* visualization of the sampled representation space.
(b) The token *human* compared with the token *cells* within a *pca* visualization of a second samples representation space.
(c) Juxtaposed coordinates are qualitatively similar across the set of probe words.

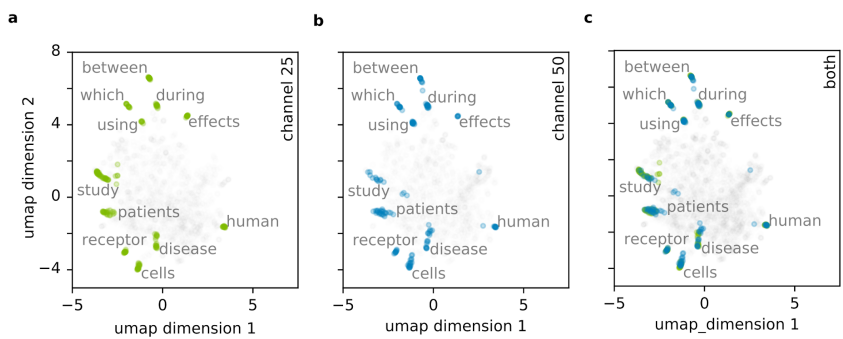

**Figure 2** Output channels are aligned, facilitating pooling
(a) Top-layer activation for probe words at output-channel 25. Vector coordinates are transformed from 768-dimensions to 2-dimensions for visualization, using umap. A random sample of token are included (gray) to reveal regions of high and low density.
(b) Top-layer activation for probe words at output-channel 50.
(c) Representations for a set of probe words are stable across two sequence positions.

### 1.3.3
*Encoder architecture*

Text was encoded using transformer network encoders (*Supplementary Figure 1*). Title and abstract text were concatenated, tokenized, and passed to the embedding layers of the encoders. Associated metadata such as the PMID identifier were not joined to the text sequence, nor encoded into semantic coordinates. Top-layer activations were sampled via mean-pooling over tokens. We also investigated other pooling approaches, including <CLS> token top-layer activation, mean-pooling over whole words, and concatenation of <CLS> activation with mean-pooled tokens. After encoding, title and abstract sequence were represented by a single 1x768 dimensional vector, recorded as a float16 array inserted into the MySQL database alongside the corresponding record.

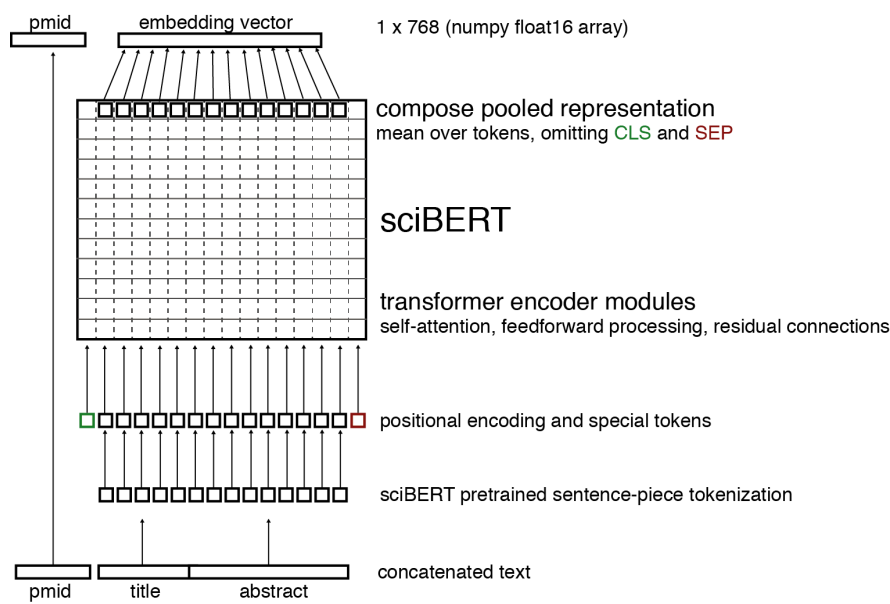

**Supplementary Figure 2** Contextual word embeddings from transformer networks. Encoding procedure for title and abstract text.

### 1.3.4
*Mean-pooled contextual word representations*

When mapping large-scale patterns in research documents, supervised labels may be unavailable or of uncertain utility. In this context, researchers may wish to use pretrained models without supervised fine-tuning. We investigate the feasibility of this approach, with attention to domain mismatch and strategy for obtaining representations.

Neuroscience is an appropriate test area for science of science, because of its interdisciplinary breadth and complexity of paradigms. Within neuroscience, we examine two nearby specialty journals that would be challenging for a non-expert to distinguish: *Journal of Neurophysiology* versus *NeuroImage*. The two journals are characterized by overlapping vocabulary, yet different scientific portfolios. For example, the first is concentrated in electrophysiology studies, while the second is concentrated in fMRI studies. The two also share a small set of topics in common, such as microscopic fluorescent-imaging. Note that *Journal of Neurophysiology* also spans a broader range of topics. On average, experts would be able to distinguish these sources easily from title and abstract alone (though we would expect some errors, for the set of topics shared between both journals). Can semantic encodings pooled from a pretrained model distinguish between these two journals, without explicitly fine-tuning using the journal labels?

Representations for entries from the two neuroscience journals were retrieved from a relational database, resulting in a matrix of size *N_samples* by *N_dimensions*. Samples were reduced in dimensionality using principle component analysis to reduce measurement artifacts from the curse of dimensionality, also demeaning the densities and recentering around zero. We examine a subspace of the full biomedical embedding for three encoder models. Results are shown below, where each row is a different encoding strategy (Figure 3).

First, the embedding space generated from probably the best known transformer encoder: the *bert-base-uncased* model, where title-abstract embeddings from the two journals occupy largely overlapping regions. The two densities exhibit peaks close to one another, reducing their discriminability. This is not unexpected, since the two journals share similar vocabularies. Yet they serve different research communities who use those vocabularies in different epistemic contexts, following different conventions.

The second embedding space is generated from a strong static word-embedding baseline, the *word2vec-pubmed* model. Here the embedding volumes are moderately less overlapping, with peaks in density somewhat less close together.

Third, the embedding space generated from the *scibert-scivocab-uncased* model. In the third row, the volumes occupied by these two journals are more distinct, with clear separation in the respective peaks in density. Here too, the larger breadth of topics covered in *Journal of Neurophysiology* is evident in the low-dimensional embeddings. Qualitatively, the third row of embeddings best differentiates between these two specialty journals. Next, we quantify these observations from the perspective of discriminability.

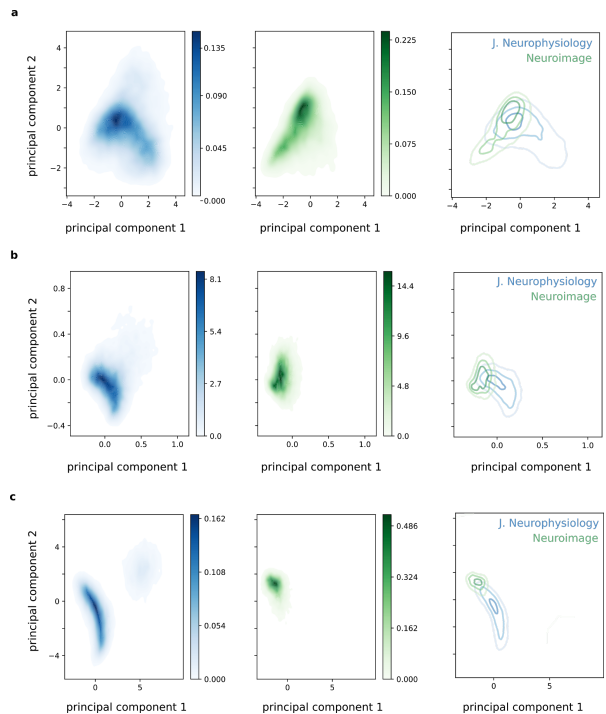

**Figure 3** Visualizing encodings of subtly distinct specialty journals
(a) In the benchmarking task, after *bert-base-uncased* encoding, distinct journals occupy overlapping regions in the low dimensional projection of the embedding space, with peaks in density close to one another.
(b) Benchmark journals are somewhat better separated after *word2vec-pubmed* encoding.
(c) Benchmark journal densities are best separated and differentiated by size after *sci-BERT* encoding.

Since *Journal of Neurophysiology* and *NeuroImage* are associated with different research communities and their respective semantic specializations, content from the two journals should be separable by an effective unsupervised encoding scheme. Given a sample article vector from a given journal, its nearest neighbors should belong to the same journal. Note that this property will not be true for any two random journals: rather, separability serves as useful benchmarking task for these two journals in particular, because *Journal of Neurophysiology* and *NeuroImage* publish in distinct niches. A more general accounting of the landscape of journals with respect to their similarities and differences is an important open problem for the science of science.

Results are summarized in Table 1 and Table 2, for two different random sample sets of 5000 title-abstract concatenations (see *Methods*). Our results highlight an important issue: powerful models borrowed from deep learning need not outperform less fashionable models, if the simpler encoders are better matched to the research question at hand. In the benchmark task, a simpler word2vec model trained on pubmed open access text (Pyysalo et al., n.d.), aggregated using mean-pooling, handily outperforms the bert-base-uncased model (correspondingly aggregated using mean-pooling of top-layer activations).

**Table 1** Compare pre-trained models (sample *A*, *n*=5000)
Comparison of informetric nearest neighbor retrieval scores for three encoders. A specialized transformer encoder more effectively distinguishes between two similar neuroscience journals, whether considering precision alone, or in conjunction with sensitivity at single or multiple scales.

|  | Precision@500 | Precision@R | MAP@R |
|---|---|---|---|
| *bert-base-uncased* | 0.83 | 0.62 | 0.70 |
| *word2vec-pubmed* | 0.87 | 0.65 | 0.75 |
| *scibert-scivocab-uncased* | **0.91** | **0.69** | **0.80** |

**Table 2** Cross-validation of retrieval performance (sample *B*, *n*=5000)
Information retrieval discriminability scores are robust to sampled cross-validation.

|  | Precision@500 | Precision@R | MAP@R |
|---|---|---|---|
| *bert-base-uncased* | 0.83 | 0.61 | 0.70 |
| *word2vec-pubmed* | 0.87 | 0.65 | 0.74 |
| *scibert-scivocab-uncased* | **0.90** | **0.68** | **0.80** |

As is clear from the premise of this paper, the strongest model by far in our benchmark task was the *scibert-scivocab-uncased* model. This transformer network encoder was trained on biomedical and computer science texts (Beltagy et al., 2019), including a custom scientific vocabulary which is better matched to the unigram statistics of biomedical text compared to *bert-base-uncased*. When other fundamental research design choices are made appropriately, deep transformer encoders are an extremely powerful option for unsupervised text analysis in biomedical applications.

In supervised settings, it is common to designate a classification token (*CLS*) and use its activation to summarize a given input sequence. Since the classification token is pretrained with the next-sentence prediction task, it may be expected to contain a coherency signal or some other semantically relevant information even without fine-tuning. We explored this option for unsupervised embeddings and compare it to two simple pooling approaches over the subword tokens themselves. For the two approaches pooling over the contextual word vectors, the *CLS* token is omitted. In the first case, we aggregate over all subwords (*tokens*). In the second case, we average only over subword tokens spanning 5 or more characters (*long tokens*).

Sampled raw embeddings from our pooling experiments are shown in Figure 4, compressed to two dimensions using *PCA*. In the top row, bert-base-uncased pooling methods are characterized by extensive overlap between samples from the two journals. Based on the leftmost column, embeddings based on the *CLS* token alone fail to separate the two specialty journals. Interestingly, *CLS* embeddings are also characterized by greater levels of discontinuity for both *bert-base-uncased* and *scibert-scivocab-uncased*. Pooling based on all subwords versus long subwords only appear qualitatively similar, with perhaps a more uniform share of variance across the first two principal dimensions in the *long tokens* scenario.

Discriminability of the two niche journals across each pooling method is presented in Table 3. Our results suggest that the *CLS* token is not an effective semantic summary of input text without additional fine-tuning. In fact, mean-pooling from *bert-base-uncased* outperforms the *scibert-scivocab-uncased CLS* vector. Embeddings based on all tokens versus only the longest tokens appear similar, with a slight performance gain from excluding short tokens. Since averaging over all subword tokens is simpler than testing the length of each token and finding a mean over that subset alone, we have continued to use the simpler pooling approach despite its slight lag in precision. However, we feel that the search for better pooling approaches merits further work, and we revisit this issue below (see *Discussion*).

**Figure 4** Pooling strategy
Sample representations from *Journal of Neurophysiology* and *NeuroImage* based on the following encoding + pooling schemes:
(a) *bert-base-uncased + [CLS]*
(b) *bert-base-uncased + mean over long subwords*
(c) *bert-base-uncased + mean over all subwords*
(d) *scibert-scivocab-uncased + [CLS]*
(e) *scibert-scivocab-uncased + mean over long subwords*
(f) *scibert-scivocab-uncased + mean over all subwords*

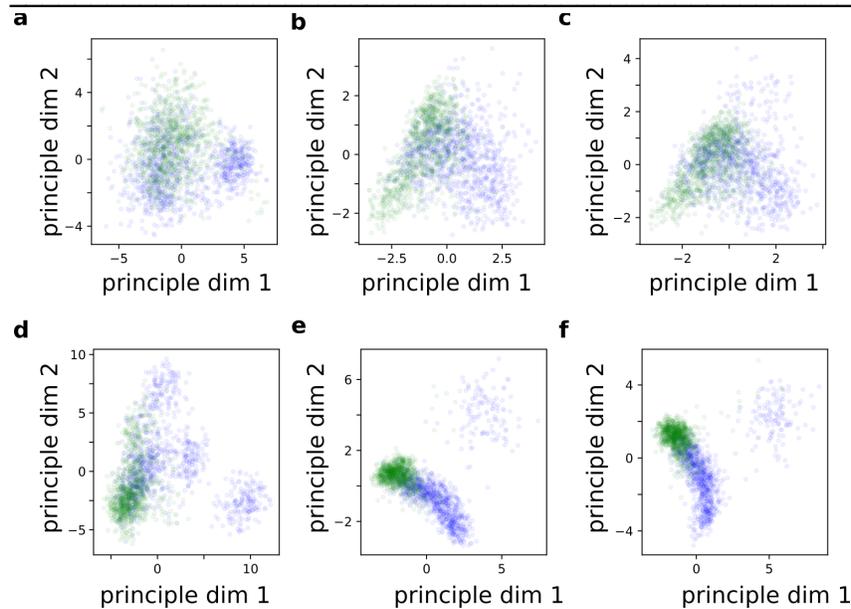

**Table 3** Pooling strategy
Comparison between pooling via the special classification token *CLS* versus mean over all other regular tokens, for a general transformer encoder compared to an encoder specialized for the domain of biomedicine. Selective averaging of long subwords can improve representations further.

|  | Precision@500 | Precision@R | MAP@R |
|---|---|---|---|
| bert *[CLS]* | 0.75 | 0.59 | 0.65 |
| scibert *[CLS]* | 0.82 | 0.63 | 0.71 |
| bert subwords | 0.83 | 0.62 | 0.70 |
| bert long subwords | 0.84 | 0.63 | 0.72 |
| scibert subwords | 0.91 | 0.69 | 0.80 |
| scibert long subwords | 0.91 | 0.70 | 0.81 |

1.3.5 *Rendering and Investigating Semantic Maps of Science*

Next, we present a suggestive application of our approach to mapping biomedical research (Figure 5), the set of all based on biomedical title-abstract entries published in 2009, from the PubMed/MEDLINE corpus. Encoded text is compressed for visualization using *umap*. Embedded title-abstracts cluster into distinct densities, which themselves are organized into larger communities. Hotspots of dense domains reveal major research communities, with shared linguistic practices. Other areas are more sparsely populated, often stretching between or budding off from dense cores.

We compare the map of published work to the map of the references they cite, including works cited multiple times. Communities receiving the most citations form a backbone, around which new publications are contextualized. Our method reveals areas that are highly cited, but less sparsely explored by published work—such as the dense cores in the upper right. Other areas are both highly cited as well as actively populated by new publications—such as the dense band in the lower left. Finally, a particularly interesting set of nuclei are densely populated by new publications, in regions that are not yet highly cited, extending the research frontier to new topics. Science is a collective search procedure, and semantic mapping can help to understand its form and dynamics.

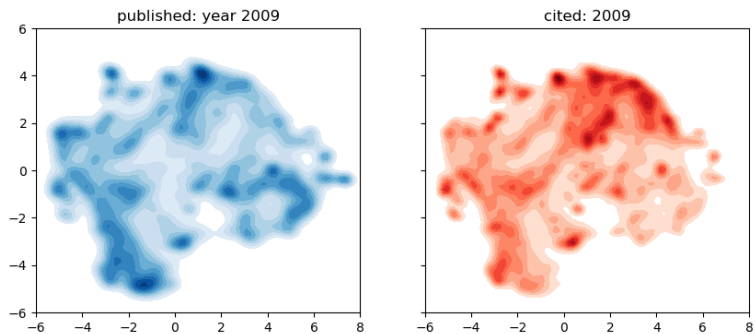

**Figure 5** The promise of deep transformer embeddings for mapping biomedical research
(a) Density map of encoded title-abstracts from 2009 publications, reduced to two dimensions with *umap*.
(b) Density map of encoded title-abstracts cited in 2009, including duplicates.

1.3.6 *Semantic Metrics of Science from Transformers*

Finally, we explore the range of scientific and information metrics that become possible with these models and pooling.

**Semantic Breadth.** Distances between elements associated with a scientific or technological work, or a collection of such works, can be measured in order to calculate how broadly or narrowly those elements are distributed in the semantic space inscribed by the pooled dimensions across channels of the transformer. Elements could be MeSH (Medical Subject Heading) keywords for PubMed-indexed paper or papers; U.S. Patent & Trademark Office codes in a U.S. patent or collection of patents; words, windows of characters (words and subwords), or entire passages such as the abstract from a research paper or papers. We can measure how far each keyword, word, character window or passage is to every other in the space, on average. When the collection is composed of multiple scientific works, we can measure how far each document is from every other document, on average. We can also take the centroid (or high-dimensional average) of a work or collection, measure the semantic distance of each element from that centroid, then take the average of those distances--the standard deviation or $\sigma$--as a measure of semantic breadth or dispersion (Shi et al., 2019). Collections can be defined for entire populations (e.g., the research papers by scientists in each European country), for institutions (e.g., papers by scientists in each university), for individuals (e.g., papers by each scientist), or more specific, custom groupings with a theoretical justification (e.g., scientists under 40...or over 80). Each of these measures could be measured in the same way for a static word embedding, as from word2vec, but the use of contextual embeddings from transformers allows us to directly embed larger passages with semantic fidelity. They also allow completely novel measurement possibilities, such as the semantic breadth (and potential ambiguity) of a single word or phrase across all of the documents in a collection.

A measure of semantic breadth could enable us to meaningfully compare teams, departments, institutions, cities, and countries that produce distinct collections of work, and explore questions like the degree to which demographic, cultural, ethnic, or experiential diversity might lead to semantically broader output (Hong & Page, 2004). They also enable us to ask questions about variations in individual and collective search in science, such as under what conditions will an scientist draw up a more or less semantically broad collection of ideas represented by the appended title and abstracts of referenced work. Given the increasing prevalence and scale of textual data from scientific and technological artifacts (Evans & Aceves, 2016; Gentzkow et al., 2019), detailed spatiotemporal analyses will also be easier to undertake, allowing us to explore breadth at each spatial and temporal cleavage point to understand scientific and technological innovation.

When we focus on a single scientific or technological work, semantic breadth may be interpreted as an indicator of novelty, because transformers are optimized to make the documents that appear frequently within their training sample highly likely—with their texts projecting to very similar spaces, as they share similar contexts. A research paper that pulls together words or passages that are diverse and unlikely to have been combined before is also novel (Uzzi et al., 2013). A threshold-based variant of this approach was recently used to measure the novelty of research in research papers during the COVID crisis, identifying an increase in the distance of paper elements (M. Liu et al., 2020). Specifically, Liu et al calculated the distance between bio-entities extracted from a paper within BioBERT (Lee et al., 2019), then counted the number of entity pairs in which their distance was in the upper 10th percentile of the pairwise distance distribution, normalized by the number of such pairs in the paper.

**Semantic Distance.** The notion of semantic breadth is relevant when our analytical focus lies on elements within the set. The same underlying measurement approach can be used when our focus is on relationships *between* different sets. In this case, we will refer to semantic distance or similarity rather than breadth. If we have at least two sets with at least one element in each, we can calculate the conceptual distance between those sets as the distance between the centroids of each. At its most basic, we can calculate the conceptual distance between two documents, which is simply measuring the semantic distance between those words. We can glimpse this potential by comparing semantic similarity to prior state of the art for measuring patent similarity. First, a researcher could trace the similarity of patents by looking at the official classification used by a patent granting agency, with patents in the same class considered more similar to each other than patents in different classes (Aharonson & Schilling, 2016; Singh & Marx, 2013). A shortcoming of this approach is that categorical measures are coarse-grained, and it is unlikely that all relevant technological characteristics are taken into account, especially as category boundaries necessarily lag technological evolution (Arts et al., 2018; Singh & Agrawal, 2011; Thompson & Melanie Fox-Kean, 2005). Second, a researcher could take two patents and measure the word overlap between them (Arts et al., 2018). Such an approach falls short, however, given its limited generalization to more than pairs of documents, failing to provide an understanding of where a patent sits relative to the body of relevant knowledge. Semantic similarity within transformers addresses these shortcomings by accessing all relevant, fine-grained information within the knowledge system and quantify precisely the location of any patent or set of patents relative to any other patent or set. Such a measure could allow us to measure semantic change over time—the distance of a collection of research articles relative to those in the part, or specifically those on which they build.

**Semantic Likelihood.** Insofar as transformers like BERT represent an advance in distributed language representations, any such representation can be turned into a classifier through inversion via Bayes rule to quantify the likelihood—or unlikelihood—that a given utterance was generated by the model (Taddy, 2015). This measurement of semantic likelihood can then be used to directly measure whether a new work of science or technology is likely or unlikely—and novel—relative to its context. For example, recent research in economic sociology uses this likelihood logic by deploying a pre-trained BERT model to explore the degree to which small versus large firms generate contextually novel ideas that foretell the future of a field from quarterly earnings calls (Vicinanza et al., 2020). Specifically, the authors calculate the perplexity, or exponentiated cross-entropy loss, for each quarterly earnings call with a BERT model tuned to contemporary or future calls. This perplexity can be understood as the inverse-likelihood that each call was generated from the milieu (the BERT model tuned to) present or future earnings calls. Semantic likelihood was calculated for more than 100,000 calls from 6,000 firms to identify a firm's "prescience"—or tendency to articulate novel ideas relative to the present, which became commonplace in the future. Prescience, in turn, was predictive of increases in annual stock market returns. Semantic likelihood can be used to directly measure the conformity or disruptive novelty of a scientific or technological work, or collection of such works, relative to any other body of work on which a contextual model is pre-trained.

**Semantic Position Along Cultural Continua.** Another novel measure can be created by tracing the location of a concept relative to a semantic dimension of interest. Static embedding algorithms like `word2vec` initially received substantial attention based on its capacity to solve analogy problems, such as "man is to woman as king is to _____" (Mikolov et al., 2013). This can be solved by performing $\overrightarrow{king} + \overrightarrow{woman} - \overrightarrow{man}$, which will return a vector closest to the vector $\overrightarrow{queen}$ on a sufficient embedding space. This suggests that $\overrightarrow{woman} - \overrightarrow{man}$ inscribes a gender vector on which $\overrightarrow{queen}$ projects positively and $\overrightarrow{king}$ projects negatively. Building on this capacity, (Caliskan et al., 2017; Kozlowski et al., 2019) proposed a method of constructing cultural dimensions such as class by taking the arithmetic mean of word vectors representing class antonyms (e.g., $\overrightarrow{rich}, \overrightarrow{affluent}$ and $\overrightarrow{poor}, \overrightarrow{impoverished}$). This approach has been widely validated and adapted (Ahn, 2019; An et al., 2018; Bodell et al., 2019; Kozlowski et al., 2019). Here we propose extending it to contextual embedding models in which dimension anchors are no longer simple word points or vectors, but rather the centroid of the cloud of points that correspond to those words across contexts (see Figure 1). For example, if one sought to discover how various topics were evaluated within a given research literature (Kang & Evans, 2020), one could anchor that dimension by calculating the difference between centroids of the vectors for words *"good" and "bad"* (or averaging that vector with others constructed from related word pairs, such as *"better-worse," "right-wrong," "satisfactory-unsatisfactory," "positive-negative," "sufficient-insufficient," "effective-ineffective," "excellent-failed," "success-failure"*). Calculating the orthogonal projection of any other document onto the resulting dimension would result in a value falling between 1.0, signifying extreme positive evaluation, and -1.0, suggesting extreme negative evaluation. Although this method was pioneered in static world embeddings, it is much more accurate with transformer embeddings because of the way in which the much larger text can be natively embedded within the space, taking not only its words, but their relationships with one another into account. This approach can be used to calculate the position of words, phrases and documents on any semantic dimension of relevance. The approach of projection to bipolar semantic dimensions can be further extended to anchoring low-dimensional subspaces with multiple concepts in which words, phrases and documents can be plotted and understood as mixtures of those concepts. This can be performed by theoretically selecting a collection of "archetypes", extremal points with known and widely shared meanings and plotting all relevant words or concepts in the subspaces defined by these extremal anchors. For example, canonical articles within a discipline can be embedded and all others can be described as mixtures of them.

## 1.4
## DISCUSSION

We have presented a semantic mapping and metrics approach that leverages the power of deep transformer encoders for natural language understanding. While transformer architectures have set new records on myriad benchmark tasks including information retrieval (Guu et al., 2020; Mike Lewis et al., 2020), practitioners in specialized domains face additional challenges in order to use these models successfully. By describing these design choices and assessing their impact, we aim to provide additional guidance on how to use transformer models for embedding applications (Arora et al., 2020), in the particular context of unsupervised textual analysis for scientific publications.

The majority of work using transformer encoders for applications in informetrics and science have employed supervised methods. The pretrained sciBERT *scibert-scivocab-uncased* model employed in this paper is trained in a self-supervised fashion (making it possible to leverage large unlabeled datasets) but typically fine-tuned in a final supervised step (Beltagy et al., 2019). Other pretrained models for science with an emphasis on biology include bioBert (Lee et al., 2019), clinicalBERT (Alsentzer et al., 2019), and bioSentVec (Q. Chen et al., 2019). A native scientific tokenization scheme also likely contributed to the success of *SciBERT* compared to models that were adapted from the baseline *BERT* vocabulary. As new progress in hyperparameter choices (Joshi et al., 2020), efficient masked language model pretraining (Clark et al., 2020), weight sharing and multi-sentence structure (Sharma et al., 2020), longer sequences (Tay et al., 2020), and contrastive objectives (Reimers & Gurevych, 2019) are translated to the domains of science studies and information retrieval (Lin et al., 2020), expect to see new state-of-the-art transformer network encoders with even more power.

However, for some research applications, supervised learning can be especially challenging, or risk distorting data. For example, fine-tuning may suppress variance along features irrelevant to the supervised task, potentially masking unexpected patterns in data. From a practical perspective, researchers may also lack labeled data along axes of interest, suggesting the value of transformer networks for characterizing and discovering useful semantic differences in the fully unsupervised setting.

Contributing to possible skepticism, a subset of early transformer encoder benchmarking triumphs around question answering and common-sense reasoning have been discounted in light of unexpected statistical shortcuts identified in benchmark datasets (Gururangan et al., 2018; Thomas McCoy et al., 2019). These new architectures are not a complete solution to the epochal problem of generalized linguistic competence. For example, transformer networks struggle with the simple-seeming logical operation of negation (Ettinger, 2020). As ever, it is important to limit our enthusiasm about new tools to a factual accounting of their actual capabilities.

Our results argue that transformer network encoders are nevertheless an extremely powerful option for unsupervised text analysis in applications to science studies and semantic indexing. Two key factors for leveraging these models effectively are domain appropriateness and effective pooling.

### 1.4.1
*Domain relevance*

The first key design choice is identifying models whose pretraining is germane to the research questions of interest. Effective transfer of pre-trained models was a key innovation that helped herald the rush of progress in natural language processing. Shared pretraining for divergent downstream tasks enabled the use of deep models with expensive one-time training costs, in terms of compute and environmental cost. Yet, even in a single language, the sequential statistics of word (co-)occurrences are variable from one domain to another, and these differences manifest in internal representations (Aharoni & Goldberg, 2020). Domain-matching is a key design consideration for contextual sentence embedding methods (Wieting et al., 2015). For example, typical internet text from Common Crawl appears quite different from the formalized conventions of biomedical research. Whenever there is no fine-tuning step to bridge potential gaps between pretraining and downstream task, this issue is particularly important.

Yet when a domain-matched pretrained model is not available, successful unsupervised applications are still possible. In these situations, an intermediate re-training procedure can be employed on the text of interest, known as "domain-adaptive pretraining". Like pretraining, which typically leverages masked-word or masked-span objectives, the domain-adaptive step does not require labeled data. Nevertheless, because of the size and complexity of these baseline models, a lot of pretraining may be required for it to achieve high performance within a nonnative domain, and researchers will do well to heed the title of a recent, influential paper on the topic: "Don't stop pretraining" (Gururangan et al., 2020).

1.4.2
*Pooling contextual word vectors*

The second key design is in choosing how to summarize the high-dimensional state of a given transformer encoder. These models typically have at least 512 output channels, each associated with a 768-dimensional (or higher) activation vector. In this work, we verified that these output channels are aligned (as expected from their tied output embeddings), and validate a simple mean-pooling approach for summarizing these high dimensional representations. Averaging over the longest tokens only (having five or more characters) improved performance further.

An alternative choice, using the classification token *CLS*, resulted in a significant deterioration of performance. This may not be obvious to potential practitioners, but it is not unexpected by virtue of the next-sentence prediction task during pretraining (which learns a coherency signal for paired text sequences demarcated with the separator *SEP* token). Moreover, the next sentence prediction task itself does not seem to be beneficial for all settings (Joshi et al., 2020), perhaps because it is too easily solvable based on unigram mismatch. Improved sentence reordering objectives are available when long-range coherency signals are important (M. Chen et al., 2019; Sharma et al., 2020).

Anecdotally, we have observed long norms for tokens like *the*, *CLS*, and *SEP*, which might be expected to be semantically uninformative. Are these components diluting the semantic signal during pooling? Are they carrying important contextual information? In supervised use cases, where a decoder can be trained to implicitly re-weight, normalize, and pool over output channels, this issue is less important. More broadly, methods which pool from alternative or additional layers (rather than from the top layer alone) may potentially offer further benefits, since the quality of semantic representations varies across layers (Voita et al., 2019). Overall, best practices for weighting and/or normalization of transformer encodings are an important area for further applied research.

1.4.3

*Semantic maps of science and the metrics they make possible*

Sometimes large-scale text analysis merely points to further questions—and that is valuable. The great strength of unsupervised approaches is their ability to organize similar text samples, particularly when semantic relationships may be obscured by surface level vocabularies (e.g., using a single word in distinct senses, a common occurrence across research communities). These new methods are one further step closer to true semantic indexing. Of course, ultimately, a true semantic map may require more than distributional word statistics alone. It may require additional competencies like world knowledge and multimodal integration—a sense of the underlying affordances, interactions, and causal relationships pointed to by the language itself. Nevertheless, by distinguishing word-senses and compositional phrases, pretrained transformer network encoders mark an important step forward towards this aspiration, and enable significant advances in metrics based on these spaces. This work offers additional evidence that deep transformers offer substantial improvement in semantic encoding compared to static word embeddings. Moreover, they articulate and conceptualize a family of metrics for exploring science, technology and culture more broadly that can rest upon these powerful models.

## 1.5
## ACKNOWLEDGEMENTS

This work benefited from the helpful suggestions of Dr. Jeff Tharsen, Dr. Kazutaka Takahashi, and Dr. Allyson Ettinger. Computational resources were coordinated through the Research Computing Cluster at the University of Chicago. This work was supported by the Air Force Office of Scientific Research #FA9550-19-1-0354 and FA9550-15-1-0162; National Science Foundation #1800956 and #1422902; a contract from the National Institutes of Health Office of Portfolio Analysis; and the UMETRICS program associated with the Institute for Research on Innovation and Science at the University of Michigan.

## 1.6

**Declaration of interest**

None